\documentclass[runningheads]{llncs}

\usepackage{textcomp}
\usepackage{amsmath}
\usepackage[T1]{fontenc}
\usepackage{multirow}
\usepackage{caption}
\usepackage{float} 
\usepackage{array}          
\usepackage{tabularx}       
\usepackage{tabu}       
\usepackage{tabularray}
\usepackage{wasysym}
\usepackage{pifont}
\usepackage{verbatim}
\usepackage{xcolor}
\usepackage{orcidlink}
\usepackage{booktabs}
\usepackage{graphicx}

\usepackage{graphicx}
\begin{document}
\title{MuCoS: Efficient Drug–Target Discovery via Multi-Context-Aware Sampling in Knowledge Graphs}

%
\titlerunning{MuCoS: Efficient Drug–Target Discovery in KGs}
%
\author{Haji Gul\inst{1a}\orcidlink{0000-0002-2227-6564} \and
 Abdul Ghani Naim\inst{1b}\orcidlink{0000-0002-7778-4961} \and Ajaz Ahmad Bhat\inst{1c}\orcidlink{0000-0002-6992-8224}
}

\authorrunning{H. Gul et al.}
%
\institute{$^1$ School of Digital Science, Universiti Brunei Darussalam, Jalan Tungku Link, Gadong BE1410, Brunei Darussalam \\
\email{($^a$23h1710, $^b$ghani.naim, $^c$ajaz.bhat)@ubd.edu.bn}\\
}
\maketitle              
\begin{abstract}
Accurate prediction of drug–target interactions is critical for accelerating drug discovery and elucidating complex biological mechanisms. In this work, we frame drug–target prediction as a link prediction task on heterogeneous biomedical knowledge graphs (KG) that integrate drugs, proteins, diseases, pathways, and other relevant entities. Conventional KG embedding methods such as TransE and ComplEx-SE are hindered by their reliance on computationally intensive negative sampling and their limited generalization to unseen drug–target pairs. To address these challenges, we propose Multi-Context-Aware Sampling (MuCoS), a novel framework that prioritizes high-density neighbours to capture salient structural patterns and integrates these with contextual embeddings derived from BERT. By unifying structural and textual modalities and selectively sampling highly informative patterns, MuCoS circumvents the need for negative sampling, significantly reducing computational overhead while enhancing predictive accuracy for novel drug–target associations and drug targets. Extensive experiments on the KEGG50k dataset demonstrate that MuCoS outperforms state-of-the-art baselines, achieving up to a 13\% improvement in mean reciprocal rank (MRR) in predicting any relation in the dataset and a 6\% improvement in dedicated drug–target relation prediction.

\end{abstract}
%
\section{Introduction}

Drug target discovery lies at the core of modern therapeutic development, enabling the identification of new biological targets, the prediction of non-target effects, and opportunities for drug repurposing --- while significantly reducing experimental costs and accelerating translational timelines \cite{sachdev2019comprehensive}. Recent `computational' advances leverage knowledge graphs (KGs) to integrate heterogeneous biomedical data (e.g., drugs, proteins, diseases, side effects, pathways) into unified networks where nodes represent entities and edges capture relationships, essentially framing discovery as a link prediction problem.

For example, KG's such as KEGG50k \cite{mohamed2019drug} and Hetionet \cite{himmelstein2017hetionet} provide comprehensive, structured representations of biological components and their intricate associations. In KEGG50k, genes (that serve as proxies for protein targets) are graph-connected through complex associations with pathways, diseases, drugs and other networks. By applying knowledge graph completion (KGC) techniques to such datasets, researchers can predict previously unobserved links between drugs and genes, ultimately accelerating drug target discovery and guiding subsequent experimental validation.

Biomedical KGC methods, however, face a critical trade-off: structural embedding methods such as ComplEx-SE \cite{mohamed2019drug} capture explicit drug-target relationships but fail to generalize to unseen entities like novel drugs due to rigid geometric constraints. Conversely, graph neural approaches like NeoDTI \cite{wan2019neodti} and Progeni \cite{liu2024progeni} integrate probabilistic reasoning with GNNs for state-of-the-art drug-target prediction but remain unevaluated on relation-centric benchmarks like KEGG50k. Furthermore, none of these methods exploit the rich textual semantics embedded in biomedical triples (e.g., "Drug X $\rightarrow$ Drug-Target-Gene $\rightarrow$ Gene Z"), which could provide inductive signals for unseen entities by contextualizing relationships beyond structural adjacency.

We posit that KEGG50k’s relational triples are inherently compatible with textual encoding strategies and therefore believe that we can leverage a language transformer model like BERT’s bidirectional attention to jointly model the explicit relationships through syntactic patterns in entity-relation-entity chains. Moreover, similar to GNNs, we propose to exploit the rich contextual information inherent in the graph’s structure, such as neighbouring entities and relations associated with a given head entity and query relation.

We therefore propose \textbf{MuCoS} (\emph{Multi-Context-Aware Sampling}), a KG completion framework that overcomes these limitations by aggregating contextual information from adjacent entities and their relationships, and then integrating this semantically enriched context into a BERT model for better prediction of relationships and entities. In doing so, MuCoS advances drug target discovery in the following key ways:

\begin{itemize}
\item \textbf{Drug–Target Relation Prediction:} By leveraging optimized neighbouring contextual information from drugs and genes (heads and tails), MuCoS outperforms traditional models in predicting general and drug–target relationships.
\item \textbf{Target-tail Prediction:} The method accurately predicts potential target tails (such as genes etc.) by incorporating contextualized structural information derived from the head entity and relationship.
\item \textbf{Efficient Multi-Context Sampling:} By prioritizing informative structural patterns through density-based sampling, MuCoS reduces computational overhead while preserving high predictive accuracy, thus accelerating the discovery pipeline.
 
\item \textbf{Elimination of Auxiliary Data Requirements:} Operating effectively without reliance on extensive entity descriptions or negative sampling, MuCoS is particularly well-suited for sparse biomedical datasets.
\end{itemize}

\section{Related Work}
Drug target discovery has been approached from multiple computational perspectives. Similarity-based methods quantify relationships by computing pairwise distances—often using Euclidean or other metric functions—between drugs and their target proteins \cite{shi2018drug}. These methods typically rely on handcrafted similarity measures to distinguish interacting pairs, laying the groundwork for early-stage target screening. Complementarily, feature-based techniques, predominantly employing support vector machines \cite{zhang2017drugrpe}, formulate the problem as a binary classification or two-class clustering task to differentiate between positive and negative drug–target associations based on engineered features.

Recent advances in graph-based methods for drug target discovery have leveraged heterogeneous networks that integrate multiple similarity metrics—such as drug–drug, target–target, and cross-modal associations—to exploit the homophily principle in biological systems \cite{ban2019nrlmfbeta}. These approaches infer missing links by modelling complex interdependencies among drugs, proteins, diseases, and pathways. In parallel, the application of embedding-based techniques has evolved considerably \cite{bordes2013transe,trouillon2016complex,yang2014distmult}. For instance, Mohamed et al. \cite{mohamed2019drug} introduced ComplEx-SE, a variant of the ComplEx KGE model that adopts a squared error-based loss for enhanced accuracy. Recent works like NeoDTI \cite{wan2019neodti} and Progeni \cite{liu2024progeni} combine graph neural networks with probabilistic reasoning to achieve state-of-the-art performance in drug–target prediction.

Despite these advances, current KGC methods still face challenges in drug target discovery. Traditional embedding models depend on static, pre-trained embeddings, which hinder their ability to generalize to novel entities and interactions in rapidly evolving biomedical data \cite{gul2024contextualized}. Text-based and large language model approaches require rich and consistent annotations—a resource often sparse in biomedical domains \cite{gul2025mucokgc}. Additionally, the reliance on extensive negative sampling during training imposes significant computational burdens, particularly for large-scale datasets. These limitations motivate us to develop MuCoS as a flexible, context-aware  and computationally efficient model that integrates both structural and textual cues to drive the discovery of new drug targets.

\section{Methodology}
MuCoS addresses two knowledge graph completion tasks: \textbf{\textit{(1) Link Prediction}} (inferring missing relations in triples like \((h, ?, t)\)) and \textbf{\textit{(2) Tail Prediction}} (identifying missing tail entities in \((h, r, ?)\)). Both tasks are divided into general and drug-target-specific subtasks to balance broad applicability with a biomedical focus. Using the full KEGG50k dataset, the general subtasks predict relations/tails across all entities and relations, while drug-target subtasks use a filtered subset to predict specific relations. This dual structure ensures versatility, supporting both domain-agnostic and biomedical-specific knowledge discovery.

\begin{figure}[H]
    \includegraphics[width=\textwidth]{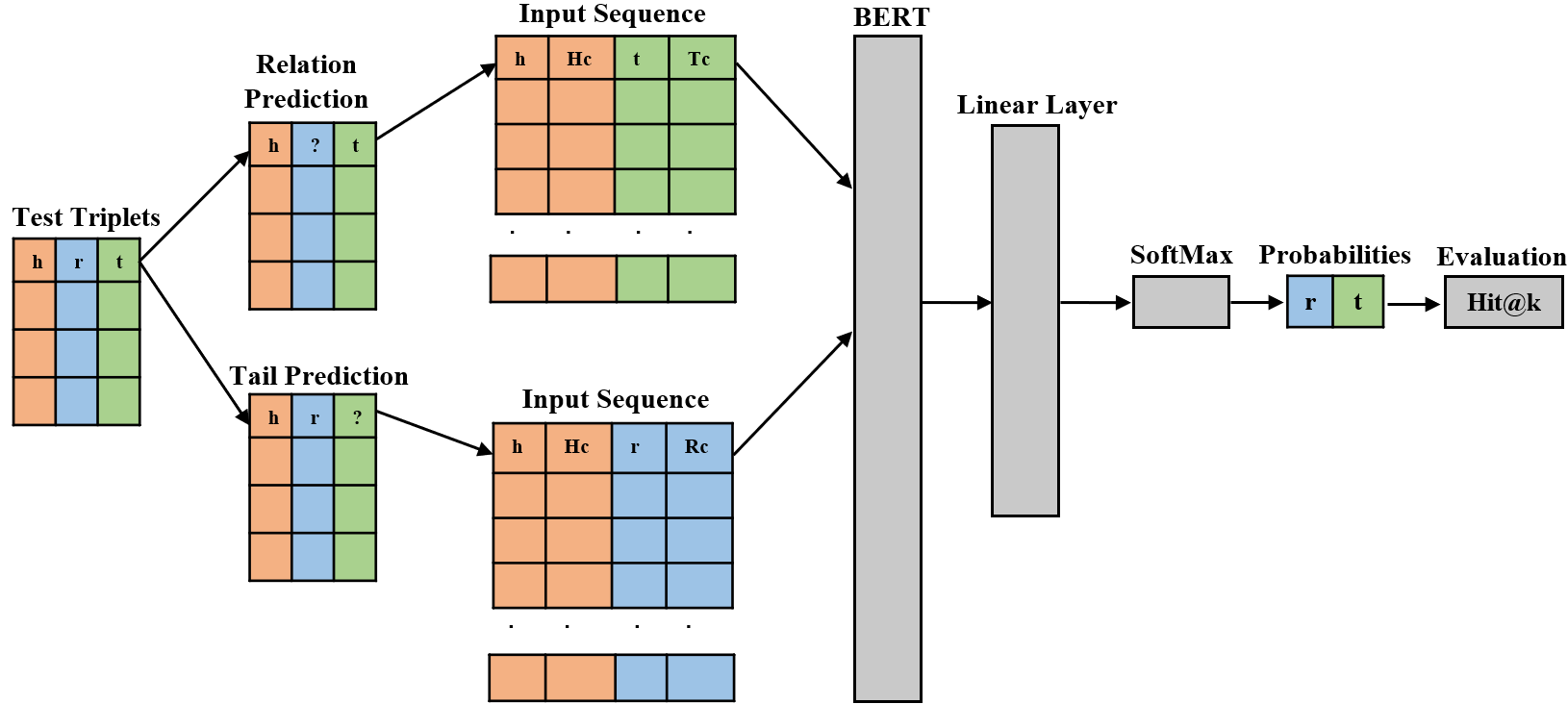}
    \caption{\scriptsize A concise overview of the MuCoS model pipeline, which is designed to predict general and drug-target relations and tail entities. The boxes on the left show the input sequence to the BERT model, where $(h)$ head, $(\mathcal{H}_c)$ head context, $(t)$ tail, $(\mathcal{T}_c)$ tail context, $(r)$ relation, and $(\mathcal{R}_c)$ relation context. This integrated context is passed through the BERT model with a linear classifier and softmax function to generate probabilities for relations and tail.}
    \label{fig:methodology}
\end{figure}

MuCoS is based on the MuCo-KGC model \cite{gul2025mucokgc}, a KGC approach that predicts missing entities ($\mathcal{E}$) in knowledge graphs by leveraging comprehensive contextual information from neighbouring entities and relations. Building on this base, MuCoS boosts computational efficiency by strategically \textit{sampling} high-density contextual information from both entity and relation-neighbouring contexts before integrating it with BERT for precise predictions. While maintaining applicability to general KGs, the sampling process enables MuCoS to accelerate KGC training on large datasets for relation and tail discovery, achieving approximately a 175-fold speed-up over MuCo-KGC in our experiments. Figure \ref{fig:methodology} provides an overview of the MuCoS pipeline. The subsequent sections detail the computations of the contextual information and the sampling process in the MuCoS pipeline.
\noindent
Given a head $(h)$, tail $(t)$, a relation $(r)$ between them, MuCoS first figures out the corresponding neighbouring contexts, i.e., the head context $(\mathcal{H}_c)$, the tail-context $(\mathcal{T}_c)$ or the relationship context $(\mathcal{R}_c)$. Based on the task at hand, relevant contexts are then concatenated and passed on to a BERT model with a linear classifier and softmax function to generate probabilities for relations or tail.

\begin{figure}[H]
    \centering
    \includegraphics[width=\textwidth]{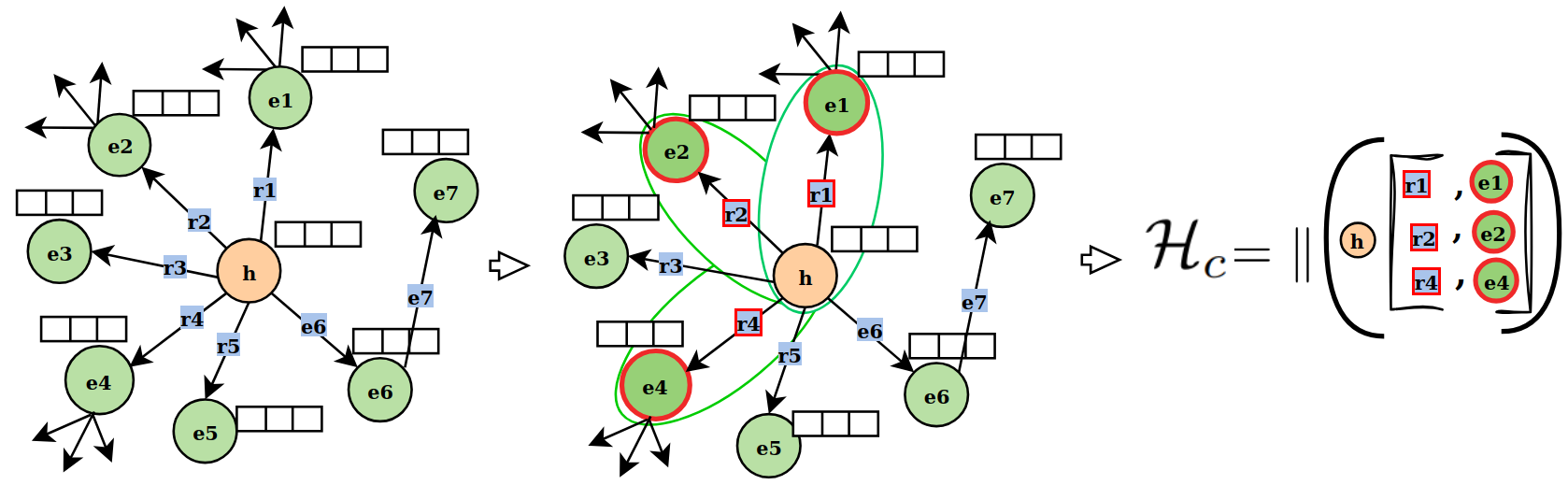}
    \caption{\scriptsize MuCoS \( \mathcal{H}_c\) construction. The left graphical view illustrates one hop head $h$ context, which consists of the set of relations \( \mathcal{R}(h) \) (\( r_1, r_2, r_3, r_4, r_5, r_6 \)) and the set of neighbouring tail entities \( \mathcal{E}(h) \) (\( e_1, e_2, e_3, e_4, e_5, e_6 \)) associated with the head entity \( h \). The middle view shows the sampling process, where only the top-\( n \) (suppose $n = 3$) tail entities $e$ are selected and concatenated ($\Vert$) based on their density \( \rho(e) \), to calculate the optimized head context  \( \mathcal{H}_c \).}
    \label{fig:dc}
\end{figure}

\textbf{\textit{Head Context $\mathcal{H}_c$ :}} To extract the contextual information for the head, i.e., $H_c$, we first identify the relations associated with the head entity $h$, i.e., the relation neighbourhood $\mathcal{R}(h)$. If $l$ relations are associated with $h$ from the set $\mathcal{R}$ of all relations $r_i$ in the graph,$G$, then:
\begin{equation}
\mathcal{R}(h) = A_{i=1}^{l} \left( \{ r_i \mid (h, r_i, e_j) \in \mathcal{T}, e_j \in \mathcal{E} \} \right)
\end{equation}
where $A(\cdot)$ is the concatenation operation $\Vert$, $\mathcal{T}$ is the set of training triples, $\mathcal{E}_t$ is the set of all tail entities, and $r_i$ represents each relation associated with $h$.
Next, we find the tail entities $e$ that are neighbours (i.e., have a direct connection) with the head entity $h$, i.e., tail neighbourhood $\mathcal{E}(h)$, using the identified relations in $\mathcal{R}(h)$. Assuming $m$ neighbour tails, $\mathcal{E}(h)$ is expressed as:
\begin{equation}
\mathcal{E}(h) = A_{i=1}^{m} \left( \{ t_i \mid (h, r_j, e_i) \in \mathcal{T}, r_j \in \mathcal{R} \} \right)
\end{equation}
\textbf{\textit{Sampling:}} While MuCo-KGC \cite{gul2025mucokgc} integrates $\mathcal{R}(h)$ and $\mathcal{E}(h)$ calculates the head context, we introduce a density-based sampling for context calculation $\mathcal{H}_c$, where the density \( \rho(e) \) of an entity \( e \in \mathcal{E}(h) \) is defined as its frequency of appearance in $\mathcal{T}$.
\begin{equation}
\rho(t) = | \{ (h,r,t) \in \mathcal{T} \}|, for ~ any ~ h, r 
\end{equation}
Using these density values, we select $n$ entities of highest density values and the relationships between the head node $h$ and these top-$n$ selected entities:
\begin{equation}
\text{top}_n(\mathcal{E}(h)) = \text{sort}(\mathcal{E}(h), \text{by } \rho(e))[:n]
\end{equation}
\begin{equation}
\mathcal{R^*}(h) = A_{i=1}^{n} \left( \{ r_i \mid (h, r_i, e_j) \in \mathcal{T}, e_j \in \text{top}_n(\mathcal{E}(h)) \} \right)
\end{equation}
The optimized head context $\mathcal H_c$ is then defined as:
\begin{equation}
\mathcal{H}_c = \mathcal{R^*}(h) \cup \text{top}_n(\mathcal{E}(h))
\end{equation}
Figure \ref{fig:dc} illustrates this sampling process, highlighting only a select subset of high-density neighbours (shown in red border) used to compute the aggregated context $\mathcal{H}_c$. We follow the same procedure to compute the tail context $\mathcal{T}_c$ (for a given tail) required along with head context $\mathcal{H}_c$ in the relation prediction task. 

\begin{figure}[H]
    \centering
    \includegraphics[width=\textwidth]{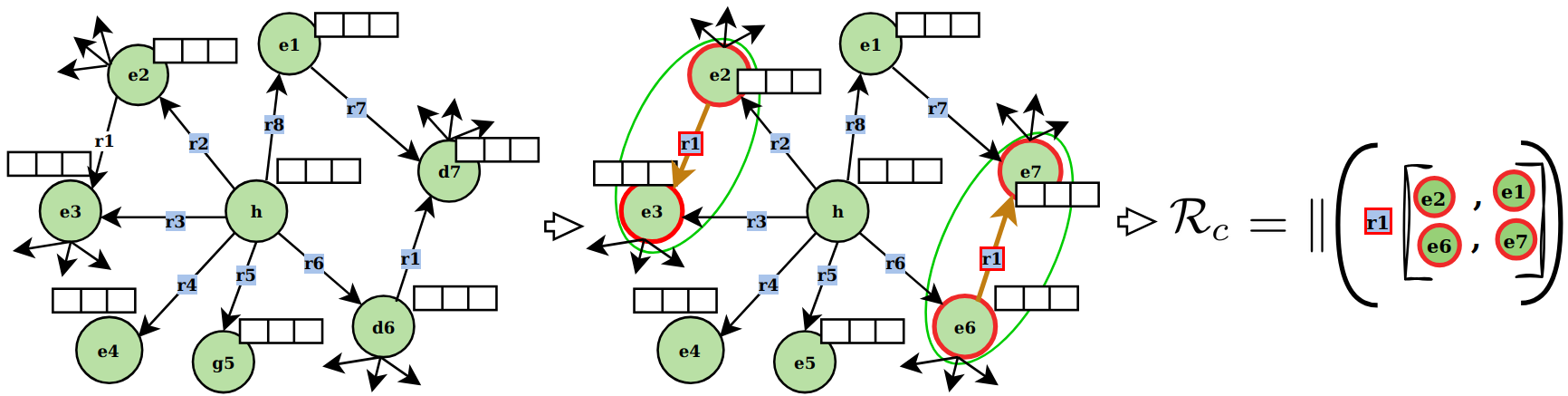}
    \caption{\scriptsize \( \mathcal{R}_c \) construction. The left view illustrates the relationship \( r_1 \) and entities (head, tail) connected by \( r_1 \). The graph in the middle depicts optimization, selecting the top \( k \) (suppose  \( k = 2 \)) entities based on density \( \rho \), retaining pairs such as \( (e_2, e_3) \) and \( (e_6, e_7) \) The optimized context \( \mathcal{R}_c \) is aggregated using concatenation ($\Vert$), as shown in the right section.}
    \label{fig:rc}
\end{figure}

\textbf{\textit{Relation Context $\mathcal{R}_c$ :}} To acquire the relation context $\mathcal{R}_c$, we identify all entities (heads and tails) associated with the operational relation $r$ in the knowledge graph $\mathcal{G}$. This includes the set of heads (e.g., drugs) $e_i$ and tails (e.g., genes) $e_j$ connected by $r$:
\begin{equation}
    \mathcal{E}(r) = A_{i,j=1}^{o}\left(\{e_i, e_j\} \mid (e_i, r, e_j) \in T\}\right)
    \label{eq:rc}
\end{equation}
\textbf{\textit{Sampling:}}
From the set of entities in $\mathcal{E}_c$, the top-$k$ elements with the highest density values  $\rho(e)$ are selected to generate the optimized relationship context $R_c$.
\begin{equation}
\mathcal{R}_c = \text{top}_k(\mathcal{E}(r)) = \text{sort}(\mathcal{E}(r), ~ ~ \text{by } (\rho(e_i)+\rho(e_j)) )[:k]
\end{equation}
$R_c$ therefore provides a focused global perspective on \( r \)'s patterns, enhancing generalization without over-exacerbating the time complexity. Figure \ref{fig:rc} depicts the sampling process involved in computing $\mathcal{R}_c$, highlighting the selection of $k$ high-density entity pairs (shown in red border) involved with the relation $r$ to form the optimized relationship context.
Following the extraction of contextual information via density-based sampling, MuCoS integrates these contexts into a BERT-based framework for prediction. The process for each subtask, leveraging the KEGG50k dataset and its filtered drug-target subset, is detailed below:

\begin{itemize}
    \item For task \textit{\textbf{(1), link prediction}}, which includes two subtasks:
    \begin{itemize}
        \item \textit{General link prediction $(h, ?, t)$ }: The concatenated representations $\mathcal{H}_c$ (head context) and $\mathcal{T}_c$ (tail context) are combined with the head entity $h$ and tail entity $t$ to form the input sequence $[h, \mathcal{H}_c, t, \mathcal{T}_c]$. This sequence passes through BERT’s transformer layers, generating a contextualized representation for each token. A classification layer then predicts the relation $r$, with a softmax function calculating the probability distribution over all relations:
        \begin{equation}
            P(r \mid h,t) = \text{softmax}(W \cdot \text{BERT}(h,H_{c},t,T_{c}))
        \end{equation}
        \item \textit{Drug-target link prediction $(h, ?, t)$}: Following Mohamed et al \cite{mohamed2019drug} in this case, we filter the dataset to consider drug-target relations only. Other than that, we follow the same methodology as above where the input sequence $[h, \mathcal{H}_c, t, \mathcal{T}_c]$ is processed by BERT to predict the drug-target-specific relations $r$.
        
    \end{itemize}
    
    \item For task \textit{\textbf{(2), tail prediction}}, which includes two subtasks:
    \begin{itemize}
        \item \textit{General tail prediction $(h, r, ?)$}: The concatenated representations $H_c$ (head context) and $R_c$ (relation context) are combined with the head entity $h$ and relation $r$ to form the input sequence $[h, H_c, r, R_c]$, using the full KEGG50k dataset. BERT processes this sequence, and a classification layer predicts the tail entity $t$:
        \begin{equation}
            P(t \mid h,r) = \text{softmax}(W \cdot \text{BERT}(h,H_c,r,R_c))
        \end{equation}
        
        \item \textit{Drug-target tail prediction $(h, r, ?)$}: Following above, we use a  filtered drug-target subset of the KEGG50k dataset, to predict the tail entity $t$.
    \end{itemize}

    \item The \textbf{\textit{cross-entropy loss}} is used to train the model by comparing predicted probability distributions with true labels for both tasks. For link prediction (general and drug-target), the loss function is defined in Equation \ref{eq:loss_l}\textbf{(a)}, where $y_i$ is the one-hot encoded true label for the relation $r_i$, and $P(r_i \mid h, t)$ is the predicted probability. For tail prediction (general and drug-target), the loss is given in Equation \ref{eq:loss_l}\textbf{(b)}, where $y_i$ represents the true label for the tail entity $t_i$, and $P(t_i \mid h, r)$ is its predicted probability. Mathematically, they are represented as:
    \begin{equation}
        \textbf{(a)} ~ ~ \mathcal{L} = - \sum_{i=1}^{N} y_i \log P(r_i \mid d, g), ~ ~ ~ ~ \textbf{(b)} ~ ~ \mathcal{L} = -\sum_{i=1}^{N} y_i \log P(g_i \mid d, r)
        \label{eq:loss_l}
    \end{equation}
\end{itemize}


\subsection{Computational Advantage of MuCoS over MuCo-KGC}
Compared to MuCo-KGC \cite{gul2025mucokgc}, MuCoS reduces computational complexity by sampling only the most significant neighbours (based on density) from the full entity and relation contexts. MuCoS employs two sampling thresholds: $n$ for the head entity context $H_c$ and $k$ for the relation context $R_c$. To compute the complexities, we first define two terms: $(i)$ the average density ($avg\_density$) as the average number of neighbours per entity in the knowledge graph, and $(ii)$ average appearance ($avg\_appearance$) of a relation $r$ in the dataset.
\begin{equation}
    avg\_density = \frac{|T|}{|E|}, ~ ~ avg\_appearance = \frac{|T|}{|R|}
\end{equation}
where $|T|$ is the total number of triples, $|E|$ entities, and $|R|$ unique relations.

\begin{itemize}
    \item For \textbf{MuCo-KGC}, the complexity of computing the head context \( H_c \) and the relation context \( R_c \) is based on full neighbourhoods without sampling. The complexity of \( H_c \) depends on the number of relations involving the head entity \( h \), denoted as \( |\mathcal{R}(h)| \), and the number of neighbouring entities \( |\mathcal{E}(h)| \), both approximated by \( avg\_density \) (see Equation \ref{eq:a1}). The complexity of \( R_c \) is determined by the number of entity pairs connected by relation \( r \), \( |\mathcal{E}(r)| \), estimated using \( avg\_appearance \) (see Equation \ref{eq:a2}). Therefore, the overall complexity for context computation in MuCo-KGC equals:
    
    \begin{equation}
        O(2 \cdot avg\_density + avg\_appearance)
    \label{eq:a3}
    \end{equation}
    where 
    \begin{equation}
        O(|H_c|) = O(|\mathcal{R}(h)| + |\mathcal{E}(h)|) = O(2 \cdot avg\_density)
    \label{eq:a1}
    \end{equation}
    and
    \begin{equation}
        O(|R_c|) = O(|\mathcal{E}(r)|) = O(avg\_appearance)
    \label{eq:a2}
    \end{equation}

    \item For \textbf{\textit{MuCoS}}, the head context $H_c$ is computed by selecting the top-$n$ high-density neighbouring entities and their corresponding relations, and the relation context $R_c$ is computed by selecting the top-$k$ high-density entity pairs. The complexity of $H_c$ is $O(n)$ for the sampled entities and $O(n)$ for the corresponding relations, and $R_c$ is $O(k)$ for the sampled entity pairs. Thus, the overall complexity for context computation in MuCoS is:

    \begin{equation}
        O(2 \cdot n + k)
    \end{equation}
    Since sampling threshold values $n$ and $k$ are much smaller than $avg\_density$ and $avg\_appearance$ in large datasets like KEGG50k, MuCoS achieves a significant reduction in computational cost compared to MuCo-KGC.
\end{itemize}


For example, in case of the KEGG50k dataset ( with \(|T| = 63,080\), \(|E| = 16,201\), \(|R| = 9\)), $   avg\_density \approx 3.895$, and $ \quad avg\_appearance \approx 7,008.89$. Therefore, the complexity of MuCo-KGC on KEGG50k dataset is:
$   O(2 \cdot 3.895 + 7,008.89) = 7,016.68 $. For MuCoS ( with \(n = 15\), \(k = 10\): the complexity is $  O(2 \cdot 15 + 10) = 40 $.
This is a speed up by a factor of:
\begin{equation}
  \frac{7,016.68}{40} \approx 175.42,
\end{equation}
 significantly lowering the computational cost compared to MuCo-KGC. The setting of the sampling sizes for the head/tail contexts $n$ at 15 and for the relations context $k$ at 10, although empirical, is based on the ablation studies on MuCo-KGC, suggesting that the head context plays a greater role than the relationship context in model performance (see Table \ref{tab:proposed_results} for detail).
\subsection{Experimental Setup}
 We evaluate MuCoS on two prediction tasks: link and tail prediction. Each task is conducted in two settings: one where entire KEGG50k dataset is considered and another where a subset containing only drug-target relations.  In \textbf{link prediction}, the model infers the missing relation in a triple $(h, ?, t)$, considering both \textit{general link prediction} (across the entire KEGG50k dataset) and \textit{drug-target link prediction} (focused on drug-target interactions). Similarly, in \textbf{tail prediction}, the model predicts the missing entity in $(h, r, ?)$, with separate evaluations for \textit{general tail prediction} and \textit{drug-target tail prediction}. This dual formulation demonstrates MuCoS's broad applicability while enabling specialized biomedical discovery. Below we provide the details of the dataset used in our experiments, the hyperparameter settings, and the evaluation criteria.

\textit{\textbf{Dataset}:} 
The proposed model was evaluated on the KEGG50k medical domain dataset, which is a curated subset of the KEGG database, specifically designed to represent drug–target interactions and associated biological entities. KEGG50k comprises of 63,080 triples that capture diverse relationships among drugs, genes, pathways, diseases, and molecular networks. These triplets are split into 57,080 training, 3,000  validation, and 3,000 testing instances (i.e. a 90:5:5 ratio split), facilitating robust evaluation of computational models in drug target discovery. Drug-target only triplet counts are 10769, 585 and 650 for the train, valid and test sets. The dataset comprises 16,201 unique entities $\mathcal{E}$ where $( \mathcal{E}_d \cup \mathcal{E}_g) \subset \mathcal{E}$ and 9 distinct types of drug-target relationships, enabling a comprehensive mapping of pharmacological interactions. Drugs in KEGG50k are derived from the KEGG drug database where as genes, representing drug target proteins, are obtained from the KEGG Gene database and serve as proxies for protein targets. 

\textit{\textbf{Hyperparameters}:} The input sequence is tokenized with a maximum length of 128 tokens. Training is conducted over 50 epochs using the AdamW optimizer with a learning rate of \(5 \times 10^{-5}\) and a batch size of 16. Experiments were performed on an NVIDIA GeForce RTX 3090 GPU with 24 GB of memory.

\textit{\textbf{Evaluation}:} Model performance is assessed using standard metrics, Mean Reciprocal Rank (MRR) and Hits@k, as defined in Equation \ref{eq:eva}, to evaluate the accuracy of general and drug-target relations and tail predictions:
\begin{equation}
\label{eq:eva}
\text{MRR} = \frac{1}{N} \sum_{i=1}^{N} \frac{1}{\text{rank}_i}, \quad \text{Hits@k} = \frac{1}{N} \sum_{i=1}^{N} \mathbf{1}(\text{rank}_i \leq k),
\end{equation}

\subsection{Results and Discussion}
\subsubsection{Link Prediction:} Table~\ref{tab:link_prediction} demonstrates that MuCoS outperforms state-of-the-art baselines on the KEGG50k dataset. It achieves an MRR of 0.65 for general link prediction across all relations, a 13\% improvement over ComplEx-SE’s 0.52, and its Hits@1 score of 0.52 exceeds ComplEx-SE’s 0.45 by 7\%. Moreover, Hits@3 and Hits@10 scores of 0.60 and 0.86 further underscore the robust ranking performance of MuCoS. Although MuCo-KGC \cite{gul2025mucokgc} achieves state-of-the-art performance, MuCoS offers a significant computational advantage, operating approximately 175 times faster with only a small reduction in accuracy.

\begin{table}
    \centering
    \caption{Relationship prediction results over the KEGG50k dataset on both general links and drug target links only. For all the metrics except for mean rank, the higher the value, the better.}
    \label{tab:link_prediction}
    \resizebox{\textwidth}{!}{%
        \begin{tabular}{l|cccc|cccc}
            \hline
            \multirow{2}{*}{Model} & \multicolumn{4}{c|}{ General link prediction} & \multicolumn{4}{c}{Drug-target link prediction} \\
            \cline{2-9}
            & MRR & Hits@1 & Hits@3 & Hits@10 & MRR & Hits@1 & Hits@3 & Hits@10 \\
            \hline
            TransE \cite{bordes2013transe} & 0.46 & 0.38 & 0.50 & 0.63 & 0.75 & 0.69 & 0.79 & 0.86 \\
            DistMult \cite{yang2014distmult} & 0.37 & 0.27 & 0.42 & 0.57 & 0.61 & 0.50 & 0.69 & 0.81 \\
            ComplEx \cite{trouillon2016complex} & 0.39 & 0.31 & 0.43 & 0.57 & 0.68 & 0.61 & 0.71 & 0.82 \\
            ComplEx-SE \cite{mohamed2019drug}
            & 0.52 & 0.45 & 0.56 & 0.68 & 0.78 & 0.73 & 0.81 & 0.88 \\
            \hline
            MuCo-KGC \cite{gul2025mucokgc} & \textbf{0.79} &\textbf{ 0.58} & \textbf{0.73} & \textbf{0.92} & \textbf{0.94} & \textbf{0.91} & \textbf{0.96 }& \textbf{1}  \\
            MuCoS ($\approx 175$ Faster) & \underline{0.65} & \underline{0.52} & \underline{0.60} & \underline{0.86} & \underline{0.84} & \underline{0.74} &\underline{0.84} & \underline{1} \\
            \hline
        \end{tabular}
    }
\end{table}

In the drug-target relationship prediction task, which focuses on identifying relationships between drugs and their target genes, MuCoS excels with an MRR of 0.84, surpassing ComplEx-SE (0.78) by 6\%. This improvement highlights the value of leveraging contextual information from head/tail entities. MuCoS’s Hits@1 score of 0.74 slightly edges out ComplEx-SE’s 0.73, while its Hits@3 score of 0.84 reflects a 3\% gain. Achieving a Hits@10 score of 1.00 (a 12\% improvement), MuCoS ranks all correct relationships within the top ten, outperforming baselines like TransE, DistMult, and ComplEx. MuCo-KGC is the most accurate at both general and drug-target predictions (e.g., MRR of 0.94 for drug-target links), but it is hard to use on a large scale because it is so hard to compute. In contrast, MuCoS offers competitive performance with a significant computational advantage, running approximately $\approx 175$ times faster than MuCo-KGC, as shown in the complexity analysis. This efficiency, without substantial loss in predictive quality, positions MuCoS as a scalable, practical solution for real-world drug discovery applications, particularly in large-scale or time-sensitive scenarios where resource optimization is critical.
\vspace{-10pt}
\subsubsection{Tail Prediction:}
Table \ref{tab:tail_prediction_updated} illustrates the results of \textbf{\textit{tail}} prediction by comparing the proposed method MuCoS with MuCo-KGC for general scenarios and drug target scenarios. MuCo-KGC (without sampling) performs better in the general scenario, achieving higher MRR and Hits@1, Hits@3, and Hits@5. MuCoS (sampling-based), on the other hand, does better for drug target scenarios, especially in Hits@10. This shows that sampling improves prediction accuracy for drug target scenarios but is slightly worse in the general scenario. The computational cost of MuCo-KGC is significantly higher than that of the MuCoS model. Furthermore, MuCoS is still performing better than the other models for predicting relationships in KEGG50k (both general and drug targets).
\vspace{-15pt}
\begin{table*}[h!]
    \centering
    \caption{Tail prediction results on the KEGG50k dataset were evaluated for both general and drug target scenarios using methods with and without sampling.}
    \label{tab:tail_prediction_updated}
    \resizebox{\textwidth}{!}{%
        \begin{tabular}{l|ccccc|ccccc}
            \hline
            \multirow{2}{*}{Model} & \multicolumn{5}{c|}{General tail prediction} & \multicolumn{5}{c}{Drug-target tail prediction} \\
            \cline{2-11}
            & MRR & Hits@1 & Hits@3 & Hits@5 & Hits@10 & MRR & Hits@1 & Hits@3 & Hits@5 & Hits@10 \\
            \hline
            MuCo-KGC & 0.39 & 0.34 & 0.521 & 0.594 & 0.718 & 0.567 & 0.457 & 0.628 & 0.681 & 0.917 \\
            MuCoS ($\approx 175$ Faster) & 0.31 & 0.215 & 0.40 & 0.49 & 0.57 & 0.442 & 0.259 & 0.46 & 0.724 & 0.868 \\
            \hline
        \end{tabular}}
\end{table*}


\begin{table*}[h!]
\centering
\captionsetup{font=small}
\caption{Tail prediction results on FB15k-237 and WN18RR datasets. The best result for each metric is in \textbf{boldface}, and the second-best is underlined.}
\label{tab:proposed_results}
\renewcommand{\arraystretch}{1.1} 
\resizebox{\textwidth}{!}{%
\begin{tabular}{lp{1.5cm}p{1.5cm}p{1.5cm}p{1.5cm}p{1.5cm}p{1.5cm}}
\toprule
\textbf{Dataset} & \multicolumn{3}{c}{\textbf{FB15k-237}} & \multicolumn{3}{c}{\textbf{WN18RR}} \\ 
\cmidrule(lr){2-4} \cmidrule(lr){5-7}
\textbf{Methods} & \textbf{MRR $\uparrow$} & \textbf{Hits@1 $\uparrow$} & \textbf{Hits@3 $\uparrow$} & \textbf{MRR $\uparrow$} & \textbf{Hits@1 $\uparrow$} & \textbf{Hits@3 $\uparrow$} \\
\midrule
MuCo-KGC (H$_c$ Only) & 0.310 & 0.263 & 0.331 & 0.420 & 0.492 & 0.556 \\  
MuCo-KGC (R$_c$ Only) & 0.280  & 0.187 & 0.255 & 0.321 & 0.345 & 0.371 \\  
MuCo-KGC &\textbf{0.350} & \textbf{0.322} & 0.399 & \textbf{0.685} & \textbf{0.637} & \textbf{0.687} \\ 
MuCoS ($\approx 175$ Faster) & \underline{0.339} & \underline{0.278} & \underline{0.335} & \underline{0.435} & \underline{0.512} & \underline{0.566} \\
\bottomrule
\end{tabular}%
}
\end{table*}

\vspace{-15pt}
\subsubsection{Evaluation on Standard KG datasets and ablation studies:}

Since MuCoS is a standard KGC method applicable to any KG, it is reasonable to evaluate its performance against common standard KG datasets such as FB15k-237 \cite{bollacker2008freebase} and WN18RR \cite{miller1995wordnet}. We also use these datasets to report the ablation studies on the MuCo-KGC model from \cite{gul2025mucokgc}, assessing the contributions of the Head Context ($H_c$) and Relation Context ($R_c$) components. Results comparing MuCo-KGC and MuCoS on the FB15k-237 and WN18RR datasets, with tail prediction are presented in Table \ref{tab:proposed_results}. MuCo-KGC \cite{gul2025mucokgc}, our earlier method, delivers strong performance with an MRR of 0.350 on FB15k-237 and 0.685 on WN18RR, surpassing many state-of-the-art models and excelling in Hits@1 (0.322 and 0.637).

In the ablation studies, the $H_c$-Only configuration, encapsulating the adjacent entities and relationships only, attains reasonable performance with an MRR of 0.310 on FB15k-237, and an MRR of 0.420 on WN18RR. The $R_c$-Only, which utilizes global relational patterns, shows inferior performance compared to $H_c$-Only, achieving an MRR of 0.280 on FB15k-237 and an MRR of 0.321 on WN18RR, highlighting the constraints of depending exclusively on global context. Evidently, each context independently contributes to enhancing MuCo-KGC's performance, although $H_c$ has a greater role in model performance.
 
 MuCoS, our proposed method, employs density-based sampling to achieve a 175-fold speed-up over MuCo-KGC, enhancing scalability for large-scale knowledge graph tasks like drug-target prediction, while maintaining competitive accuracy with an MRR of 0.339 on FB15k-237 and 0.435 on WN18RR, alongside solid Hits@1 (0.278 on FB15k-237 and 0.512 on WN18RR) and Hits@3 (0.335 on FB15k-237 and 0.566 on WN18RR) scores; though slightly below MuCo-KGC’s peak performance. 
\section{Conclusion}
The study introduces MuCoS, a multi-context-aware sampling method that uses BERT to improve drug-target relation predictions and tail entity predictions in biomedical knowledge graphs. MuCoS employs a dual strategy—combining transformer-based textual modelling with context-aware sampling—to overcome limitations of existing models, such as poor generalization, negative sampling, and the need for descriptive entity information. It extracts and optimizes contextualized information from the head, tail, and relation entities using density-based sampling and its lexical semantics,  capturing richer structural patterns and reducing computational complexity. Experimental results show superior performance over state-of-the-art models, with improvements in MRR and Hits@1 for general and drug-target relationship prediction.

\end{document}